%% file: root.tex
\documentclass[letterpaper, 10 pt, conference]{ieeeconf}  % Comment this 
\IEEEoverridecommandlockouts                              % This command is only needed if 
\usepackage{microtype}
\usepackage{graphicx}
\usepackage{subfigure}
\usepackage{booktabs} % for professional tables
\usepackage{amsmath}
\usepackage{graphicx}
\usepackage{algorithm2e}
\usepackage{algpseudocode}
\usepackage{listings}
\usepackage{xcolor}
\usepackage{caption}
\usepackage{amsfonts}
\usepackage{svg}
\usepackage{url}
\usepackage{hyperref}
\usepackage[noadjust]{cite}

\title{ \bf \LARGE
Traffic-Aware Autonomous Driving \\ with Differentiable Traffic Simulation
}

\author{Laura Zheng, Sanghyun Son, and Ming C. Lin \\
    \thanks{The authors are with 
 Department of Computer Science, 
         University of Maryland at College Park, MD, U.S.A.
        E-mail: \{lyzheng,shh1295,lin\}@umd.edu} %
       \href{https://gamma.umd.edu/trafficdriving}{\texttt{gamma.umd.edu/trafficdriving}}
       }
       
\begin{document}

\maketitle
\thispagestyle{empty}
\pagestyle{empty}

%%%%%%%%%%%%%%%%%%%%%%%%%%%%%%%%%%%%%%%%%%%%%%%%%%%%%%%%%%%%%%%%%%%%%%%%%%%%%%%%

\begin{abstract}
While there have been advancements in autonomous driving control and traffic simulation, there have been little to no works exploring their unification with deep learning. Works in both areas seem to focus on entirely different exclusive problems, yet traffic and driving are inherently related in the real world. In this paper, we present Traffic-Aware Autonomous Driving (TrAAD), a generalizable distillation-style method for traffic-informed imitation learning that directly optimizes for faster traffic flow and lower energy consumption. TrAAD focuses on the supervision of speed control in imitation learning systems, as most driving research focuses on perception and steering. Moreover, our method addresses the lack of co-simulation between traffic and driving simulators and provides a basis for directly involving traffic simulation with autonomous driving in future work. 
Our results show that, with information from traffic simulation involved in the supervision of imitation learning methods, an autonomous vehicle can learn how to accelerate in a fashion that is beneficial for traffic flow and overall energy consumption for all nearby vehicles. 

\end{abstract}

\input{sections/intro}
\input{sections/related}

\input{sections/background}
\input{sections/method}

\input{sections/results}
\input{sections/conclusion}

% References are important to the reader; therefore, each citation must be complete and correct. If at all possible, references should be commonly available publications.

\newpage

\bibliographystyle{IEEEtran}
\bibliography{ref}

\newpage

\input{sections/appendix}

\end{document}

%% file: sections/intro.tex
\section{INTRODUCTION}
The ideal autonomous vehicle (AV) should be able to minimize the travel time of a route, maximize the energy efficiency of the vehicle, and provide a smooth and safe experience for the riders. These objectives are not only important to the passenger’s experience, but also for greater global benefits.  Annually in the US, traffic congestion accounts for 29 billion dollars in costs~\cite{emissions}, transportation accounts for 27\% of carbon emissions~\cite{EPA_emissions}, and motor vehicle accidents are the leading cause of unnatural death as of 2022~\cite{CDC_cause_of_death}. From a learning perspective, the extent an AV can improve on these objectives is affected by the environment around it, \textit{e.g.} the surrounding vehicle traffic and the road networks. 
Similarly to other multi-agent problems, improving traffic flow and energy efficiency for all vehicles in the system can also benefit each vehicle’s individual objectives. 
Inversely, the vehicle’s motion also impacts the traffic around it. An AV influences the flow of human-driven traffic by acting as a pace car to those behind it. A ``pace car'' is commonly used in racing to control the speeds of competing vehicles; this same notion can be applied to traffic control in the physical world. 
In a driving policy, we can emulate this effect by defining an individual vehicle's objective in terms of global metrics.
% flow, fuel consumption, and changes in speed. 

% Therefore, we define these objectives by taking into account the average vehicle velocity in the road network that the autonomous vehicle is traveling on (traffic flow), fuel consumption, and variations in speed over time. 

\begin{figure}[th]
\vspace*{-1.5em}
  \includegraphics[width=0.48\textwidth, trim=0cm 0cm 6cm 0cm,clip]{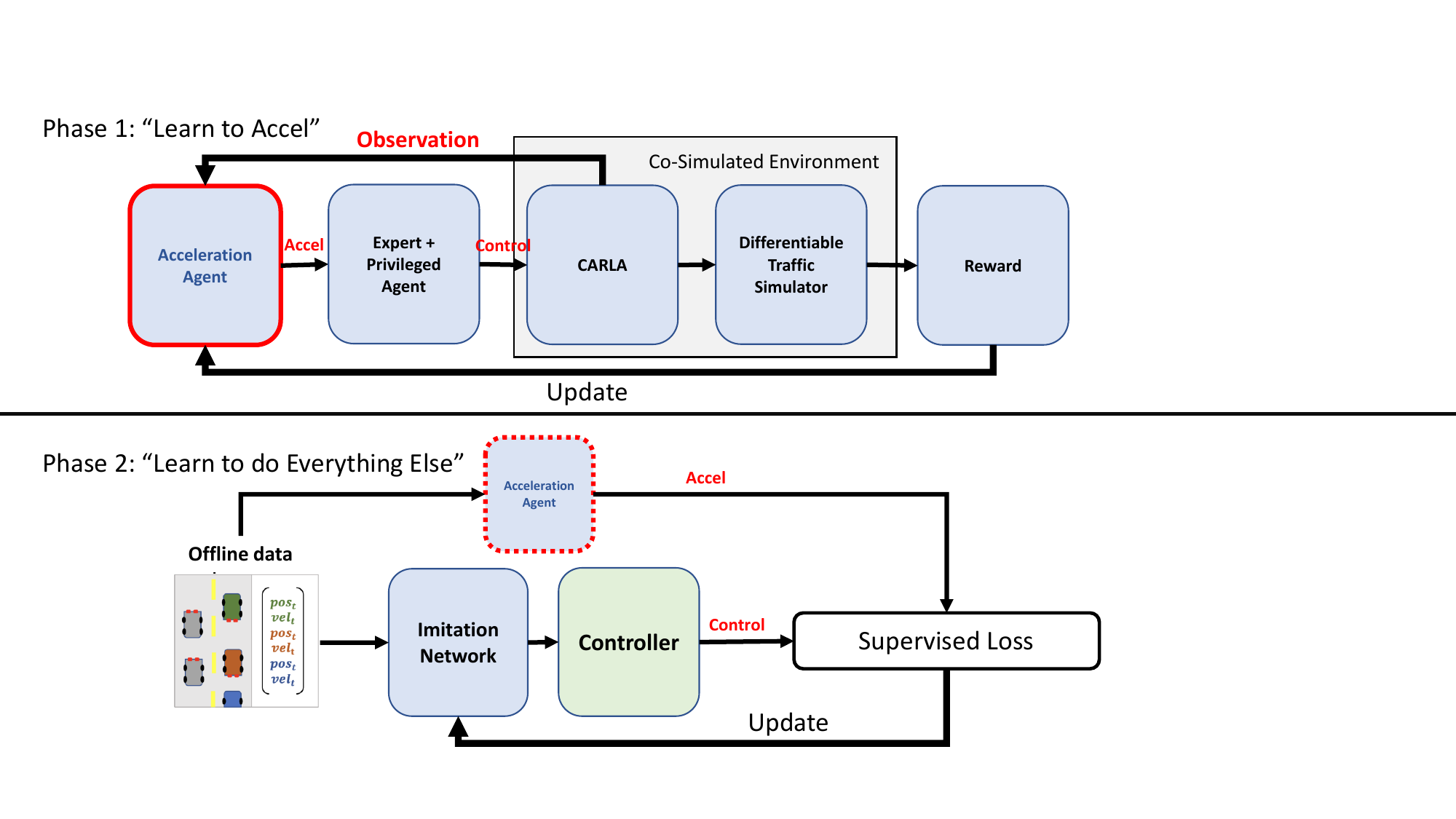}
% \vspace*{-1.5em}
  \caption{\textbf{Training for Traffic-aware Autonomous Driving.} Our method can be adapted to most existing imitation learning frameworks for driving by just adding an additional phase of training, where we isolate ``Learn to Accelerate''. 
  An agent whose action space only spans possible acceleration actions navigates through a co-simulated environment and is rewarded when it improves overall traffic flow and fuel consumption. 
  In Phase 2, where we ``Learn to do Everything Else'', the acceleration agent is frozen and supervises speed control of an imitation learning agent via distillation. 
}
  \vspace*{-2em}
  \label{fig:method}
\end{figure}

As long as a model of the traffic and its dynamics on a road network are accessible, an autonomous driving policy can directly optimize for {\em traffic flow, energy efficiency, and smooth acceleration}. 
Decades of traffic engineering research present sophisticated mathematical equations modeling traffic, including car-following models~\cite{Treiber_Hennecke_Helbing_2000, Newell_2002, Wiedemann_1974, krauss}, traffic flow models~\cite{lighthill1955kinematic, gazis1961nonlinear, aw2000resurrection, zhang2002non} and theory~\cite{Kerner}. These mathematical models, though often too simplified to account for the uncertainty of driver behaviors, are computationally efficient, differentiable, and do not require data. One possibility is to model the traffic environment with a neural network, as in some of the latest work~\cite{Olayode_Tartibu_Okwu_2021, Diehl_Brunner_Le_Knoll_2019, Jiang_Luo_2022, Do_Taherifar_Vu_2019}. Although these methods are more accurate than ODEs, they also require large amounts of data to generalize and are subject to problems associated with deep learning, such as bias and distributional shifts. In this paper, we couple a learning-based traffic control algorithm with {\em differentiable} ODE traffic models.
We use the gradients of forward traffic dynamics to guide the learning of a driving policy, so long as the policy has access to traffic information. Minimal traffic information involves simulation states of position and velocity over time, which is further explained in Section~\ref{subsec:sim-definitions}.

We present a generalizable method for {\em traffic-aware autonomous driving} (TrAAD) which takes advantage of {\em differentiable traffic simulation}. By coupling a driving environment with traffic simulation, the driving policy can retrieve traffic information during training and learn behaviors that are both beneficial to individual and global goals. 

In TrAAD, we add a phase of training in addition to traditional imitation learning for driving, where the vehicle “learns to accelerate”. This phase involves maximizing the overall traffic flow of a vehicle’s local lane, minimizing the fuel consumption of all vehicles, and discouraging the acceleration actions from being too jerky. 
Because our method supervises acceleration via distillation, it is generalizable to nearly any standard imitation learning framework, regardless of architecture or design. Our results show that our method, when implemented on top of existing state-of-the-art driving frameworks, improves traffic flow, minimizes energy consumption for the AV, and enhances the passenger's ride experience.

In summary, we present the following key contributions: 

\begin{enumerate}
    \item A simulated traffic-annotated driving dataset for imitation learning for self-driving cars; 
    \item Use of gradients from differentiable traffic simulation to improve sample efficiency for
    autonomous vehicles; 
    \item A generalizable method for traffic-aware autonomous driving, which learns to control the vehicle via rewards based on societal traffic-based objectives.
\end{enumerate}
Additional results, materials, code, datasets, and information can be found on our project website.  % at \url{https://gamma.umd.edu/trafficdriving}.

%% file: sections/related.tex
\section{RELATED WORKS} 
% TODO: I plan to add more related works + citations related to imitation learning and traffic simulation in general. 

\subsection{Autonomous Driving with Traffic Information}
Zhu et al. recently proposed a method for safe, efficient, and comfortable velocity control using RL~\cite{Zhu_Wang_Pu_Hu_Wang_Ke_2020}. Similarly to one of our objectives, they aim to learn acceleration behavior that exceeds the safety and comfort of human expert drivers. One major difference is that our work complements existing end-to-end autonomous driving systems with multi-modal sensor data, and learned acceleration behavior cooperates with learned control behavior from imitation learning rather than learning acceleration in a pure traffic simulation setting. In addition, our objective is to directly optimize on an entire traffic state, not just the objectives for the autonomous vehicle itself. The reward objectives of ~\cite{Zhu_Wang_Pu_Hu_Wang_Ke_2020} are also inferred from a partially-observed point of view. Other works have considered learning driving behavior with passenger comfort and safety in mind, but many do not directly involve traffic state information beyond partially-observed settings~\cite{Shen_Zhang_Ouyang_Li_Raksincharoensak_2020, Zhu_Wang_Hu_2020, Li_Yang_Li_Qu_Lyu_Li_2022}. Wegener et al. present a method for energy-efficient urban driving via RL~\cite{Wegener_Koch_Eisenbarth_Andert_2021} in a partially-observed setting purely in traffic simulation, however, does not address integration with current works for more complex vehicle control. In short, our method addresses a broader method for learning a policy beneficial to both individual and societal traffic objectives, while can be easily integrated into existing state-of-the-art end-to-end driving control methods. 
\vspace*{-0.5em}
\subsection{Differentiable Microscopic Traffic Simulation}
While differentiable physics simulation has been gaining popularity in recent years, differentiable traffic simulation is  under-explored, especially in applications for autonomous driving.
In 2021, Andelfinger first introduced the potential of differentiable agent-based traffic simulation, as well as techniques to address discontinuities of control flow~\cite{Andelfinger_2021}. In his work, Andelfinger highlights continuous solutions for discrete or discontinuous operations such as conditional branching, iteration, time-dependent behavior, or stochasticity in forward simulation, ultimately enabling the use of automatic differentiation (autodiff) libraries for applications such as traffic light control. One key difference between our work and ~\cite{Andelfinger_2021} is that our implementation of differentiable simulation accounts for learning agents acting independently from agents following a car-following model, and is compatible with existing learning frameworks. In addition, we optimize traffic-related learning by defining analytical gradients rather than relying solely on auto-differentiation.
Most recently, Son et al. proposed a novel differentiable hybrid traffic simulator that computes gradients for both macroscopic, or fluid-like, representations and agent-based microscopic representations, as well as the transitions between them~\cite{son2022traffic}. In our work, we focus solely on microscopic agent-based simulation to maintain relevance to autonomous driving frameworks. 
% There are no other peer-reviewed works in differentiable traffic simulation, to the best of our knowledge.

\vspace*{-0.21em}
\subsection{Deep Learning with Traffic Simulation}
Deep reinforcement learning has been used to address futuristic and complex problems for control of autonomous vehicles in traffic. One survey on Deep RL for motion planning for autonomous vehicles by Aradi~\cite{9210154} delineates challenges facing the application of DRL to traffic problems, one of which is the long and potentially unsuccessful learning process. This has been addressed in several ways through curriculum learning~\cite{Qiao_Muelling_Dolan_Palanisamy_Mudalige_2018, Bouton_Nakhaei_Fujimura_Kochenderfer_2019, Kaushik_Prasad_Krishna_Ravindran_2018}, adversarial learning~\cite{Ferdowsi_Challita_Saad_Mandayam_2018, Ma_Driggs-Campbell_Kochenderfer_2018}, or model-based action choice. In our work, we address this issue via sample enhancement for on-policy deep reinforcement learning. With differentiable traffic simulation and access to gradients of reward with respect to policy action, we can artificially generate ``helpful'' samples during learning with respect to reward. 
``FLOW'' by Wu et al. ~\cite{wu2017flow} presents a deep reinforcement learning (DRL) benchmarking framework, built on the popular microscopic traffic simulator SUMO~\cite{SUMO2018}. Wu et al. provide motivation for integrating traffic dynamics into autonomous driving objectives with DRL, defining the problem/task as ``mixed autonomy".  Novel objectives for driving include reducing congestion, carbon emissions, and other societal costs; these are all in futuristic anticipation of mixed autonomy traffic.  Based on FLOW, Vinitsky et al. published a series of benchmarks highlighting 4 main scenarios regarding traffic light control, bottleneck throughput, optimizing intersection capacity, and controlling merge on-ramp shock waves~\cite{pmlr-v87-vinitsky18a}.  We extend the environments from FLOW's DRL framework to be {\em differentiable} and show benchmark results for enhanced DRL algorithms utilizing traffic flow gradients for optimization.

%% file: sections/background.tex
\section{Background}
\label{sec:difftraffic}
% In this section, we establish formal notation and definitions that will be used throughout the rest of the paper. 
\vspace*{-0.5em}
\subsection{Simulation-related Notation and Definitions}
\label{subsec:sim-definitions}
To integrate traffic simulation into learning and optimization frameworks for autonomous driving, we need differentiable forward simulation. Agent-based traffic simulation is governed by ordinary differential equations (ODEs) known as car-following models. These ODEs describe the position and velocity of each individual agent over time. In the context of traffic simulation, the agent directly in front is considered in calculating the position and velocity. Since the position and velocity of each individual agent can be used to describe the current state of the entire simulation, we define the \textit{agent state $q^t_n$} and the \textit{simulation state $s^t$} at time step $t$ for vehicle $n$ out of $N$ total vehicles to be 
\vspace*{-0.5em}

\begin{equation}
\label{eq:sim_state}
    q_n^t = 
    \begin{bmatrix}
    x_n^t \\
    v_n^t
    \end{bmatrix}^\top,  \\
    s^t = 
    \begin{bmatrix}
    q_1^t \ldots q_N^t
    \end{bmatrix}^\top \\
\end{equation}

\subsection{The Intelligent Driver Model (IDM)}
% Before getting into the derivatives, it would be helpful to introduce the car-following ODE itself. 
% IDM was first proposed by Treiber, Hennecke, and Helbing in 2000. It 
We model traffic dynamics of human drivers in our experiments with the Intelligent Driver Model (IDM)~\cite{Treiber_Hennecke_Helbing_2000}. IDM describes how the position and velocity of each individual vehicle change over time. Intuitively, vehicles following the IDM will increase acceleration when the headway to the vehicle in front is large, and decelerate comfortably based on a maximum deceleration parameter. 
Unlike previous works in traffic simulation, we analytically derive the gradient of forward simulation to optimize the computation of backpropagation, rather than relying on auto-differentiation. 
We leave the formal definition, speedup analysis, variable definitions, and the full derivation of the analytical IDM Jacobian in the appendix, which can be found on the project website. 

%% file: sections/method.tex
\begin{figure}[th]
% \vspace*{-1em}
  \includegraphics[width=0.48\textwidth,height=6cm]{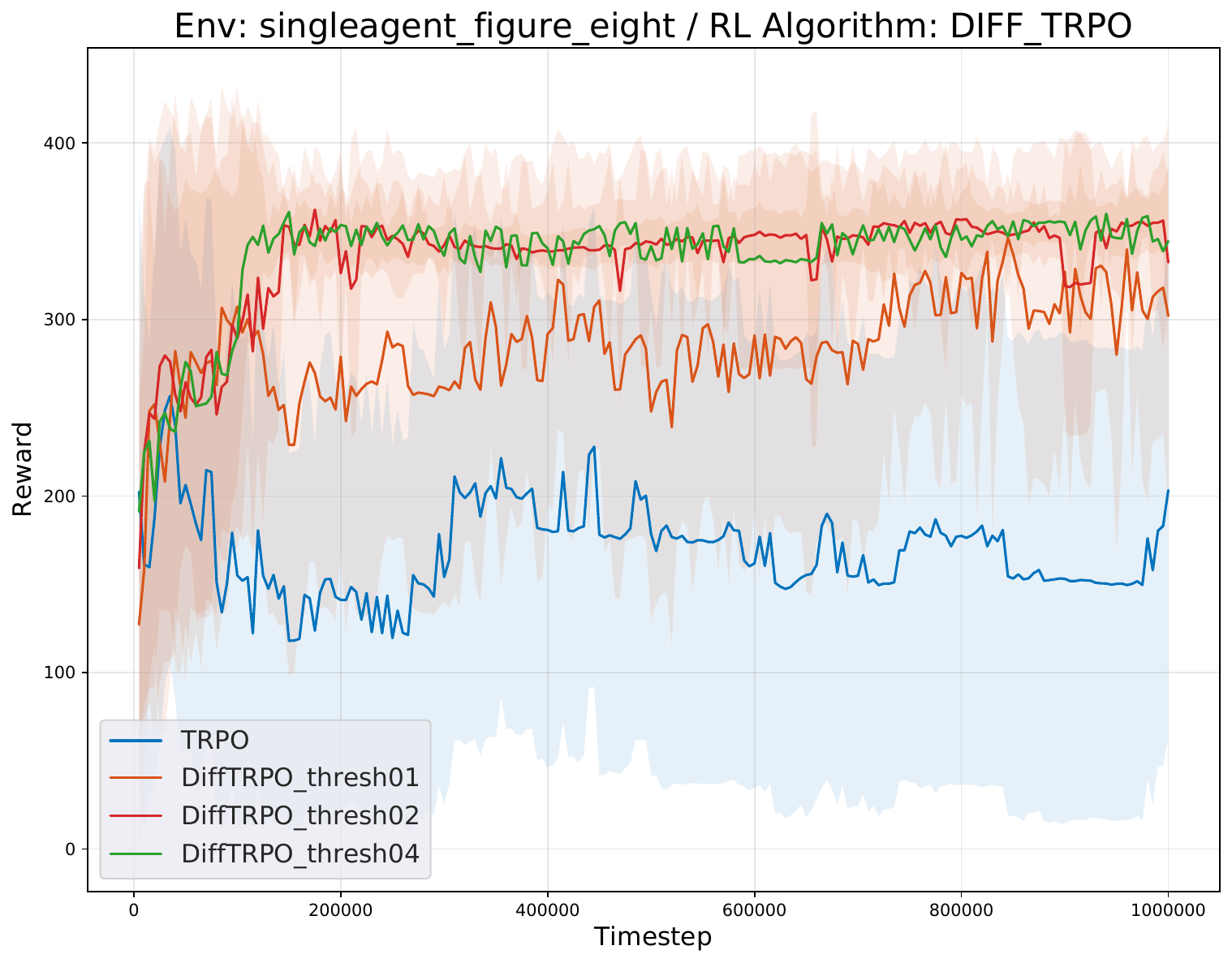}
  \caption{\textbf{Effect of Gradients for On-Policy Learning.} We show the training curve for our sample efficiency enhancement method DiffTRPO versus the on-policy algorithm TRPO~\cite{Schulman_Levine_Abbeel_Jordan_Moritz_07}, as well as various experiments varying the perturbation threshold parameter $\delta$ discussed in section~\ref{sec:sample_enhancement}. Curves are averaged over 10 runs each to account for stochasticity. A perturbation threshold $\delta$ of 0.2 and 0.4, shown in red and green respectively, yields {\em faster learning} and {\em higher reward} than the baseline TRPO in blue. Additional results for PPO are shown in supplemental materials on the project website.}
  \label{fig:grad_rl}
\vspace*{-2em}
\end{figure}

\section{Traffic-Aware Autonomous Driving}
Most, if not all self-driving models focus on replicating expert human behavior as ground truth. Additionally, FLOW~\cite{wu2017flow} introduces the vision of mixed autonomy, where AVs can drive in a way that benefits all members of the simulation. This behavior is not necessarily observed in the real world or known to humans, and thus ground truth for optimal driving for societal good is not known. Currently, there is no existing research on integrating such behavior into single-vehicle driving control. 
We present TrAAD as a practical application of the ideas presented by FLOW to current imitative autonomous driving methods. 

Differentiable traffic simulation can be directly applied to supervised learning due to the availability of simulation gradients for backpropagation. In our method, we use traffic simulation gradients to improve sample efficiency in on-policy RL. 

\textbf{Dataset Collection.}
We collect our own driving dataset in the same format as ``Learning by Cheating'' (LBC)~\cite{chen2019lbc}. Driving data is collected by an expert driver, CARLA autopilot, in CARLA Simulator~\cite{Dosovitskiy17} at a rate of 2 Hz, or every 10 frames at 20 frames per second. The expert driver is \textit{not} a learned agent and uses route waypoints and simulation information to drive optimally. 
Training data comprises 50 routes over towns 1, 3, 5, and 6, while test data comprises 26 routes. These towns describe a small suburban town, a large complex city, a square-grid city, and a highway environment respectively. Sensors include a top-down birds-eye segmentation map and three RGB cameras (left, right, and center dashcam views). Annotations include vehicle position, vehicle control commands, and traffic state of the current vehicle lane in the same format as Equation~\ref{eq:sim_state}. 
% Traffic state annotations require CARLA simulation to be synchronized with SUMO simulation. Thus, data collection runs in co-simulation with a SUMO server, whose sole purpose is to provide such state annotations for the vehicle's current lane. 
The dataset and the code to generate it can be found on our project website. 

\begin{figure*}[ht!]
\vspace*{-0.25em}
  \centering
  \includegraphics[width=0.75\textwidth,height=3.5cm]{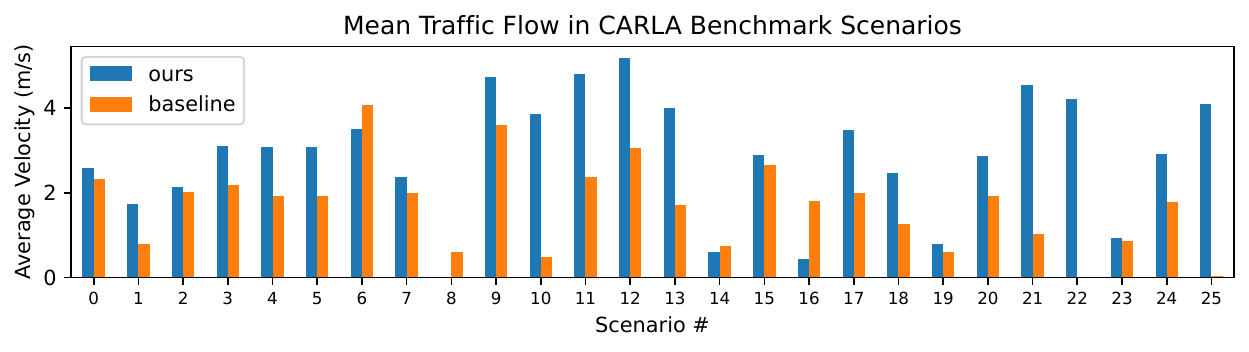}
  \includegraphics[width=0.75\textwidth,height=3.5cm]{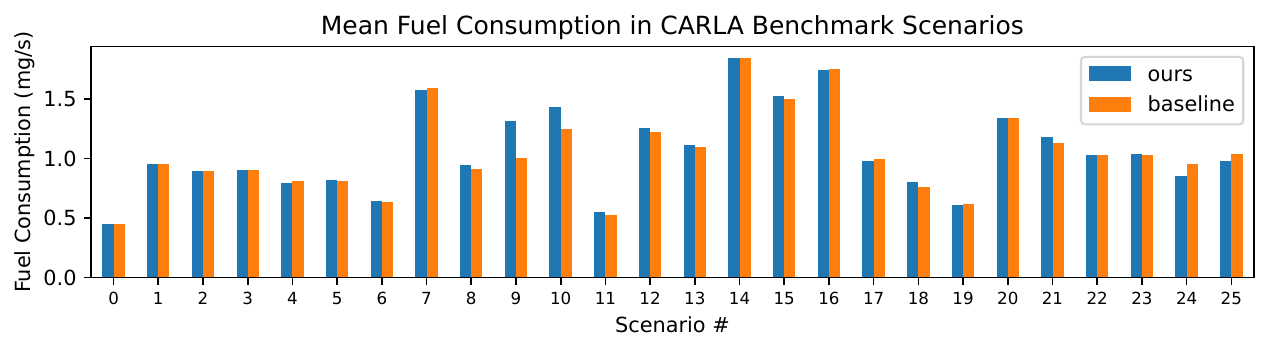}
  % \vspace*{-0.25em}
  \caption{\textbf{Comparison of overall traffic flow and fuel consumption between our method and baseline, for each CARLA test scenario.} 
  Our model consistently improves overall traffic flow over measured baseline traffic flow in nearly all scenarios, showing that our single-vehicle control can influence the traffic flow around it. Our method is also able to maintain similar, if not better, fuel consumption -- in spite of a direct relationship between increased flow and increased fuel usage.} 
  \vspace*{-1em}
  \label{fig:barchart}
\end{figure*}

\textbf{Hardware specs and Software.}
All experiments are trained on a single NVIDIA RTX A5000 GPU, Intel(R) Xeon(R) W-2255 CPU (20 cores), and 16 GB RAM, with the exception of experiments on TransFuser, which was trained on eight A5000 GPUs in parallel. 

\subsection{Phase 1: Learning Acceleration with Online RL}
The goal of this phase is to learn acceleration behavior with on-policy reinforcement learning (RL) algorithms, given that other controls such as steering are optimal via the expert driver. For our experiments, we use PPO as well as the gradient-enhanced PPO later described in Section~\ref{sec:sample_enhancement}, which improves sample efficiency by taking advantage of simulation gradients. 

\begin{figure}[htb!]
  \centering
  \includegraphics[width=8cm,height=3cm]{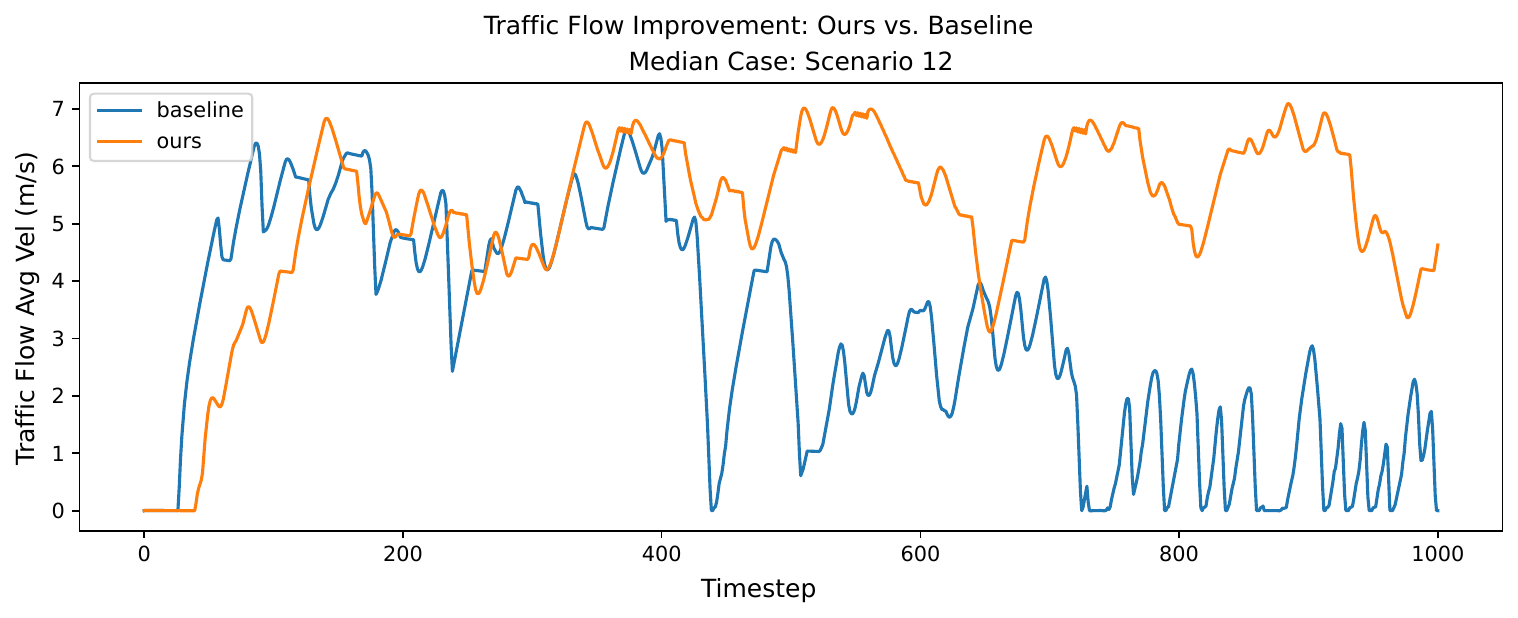}
    \caption{\textbf{Comparison of Traffic Flow Over Time in the median case.} We visualize the average velocity of traffic flow over time in the learned agent's lane for TrAAD (in orange) versus the baseline (in blue).
    TrAAD achieves {\it +64.27\%} on average.
    % Plots for all test scenarios can be found on the project website. 
    }
    \vspace*{-2em}
  \label{fig:timeseries_leaderboard}
\end{figure}

The action space of this phase is continuous and defined by a maximum acceleration $\alpha_{max}$ and minimum acceleration $\alpha_{min}$, which represent units of meters per second squared ($m/s^2$). The observation space is multi-modal and depends on the inputs of the imitation learning framework. For LBC, we collect a top-down segmentation map $M$ plus the traffic state $s$ given by the simulator.
To account for traffic laws and discourage blind acceleration, the predicted acceleration is only used when the expert determines a non-braking state. 
Reward functions from FLOW~\cite{wu2017flow} can now be used with auto-differentiation since traffic simulation is differentiable.
% It is also worth noting that reward functions from FLOW are not inherently differentiable due to the lack of differentiable traffic simulation; thus differentiable simulation enables differentiable rewards as well. 

In our experiments, we use a weighted sum of two global objectives, $R_{vel}$ and $R_{mpg}$, plus an individual jerkiness constraint for the AV.
The first term relates to the average velocity of the traffic state, the second term is a miles-per-gallon metric that relates to energy efficiency, and the third term is an L2 constraint that discourages the next predicted actions to be too far from the last predicted action for smoother acceleration. Intuitively, a passenger will feel safer if a driver slowly presses down the gas pedal to accelerate. The third term can also be viewed as a form of time-dependent regularization on sequences of actions.
The reward function $R_{comb}$ is described below, where $g^i_t$ is gallons of fuel consumed per second and $v^i_t$ represents the velocity in meters per second by vehicle $i$ at timestep $t$ after action $a_t$: 
\vspace*{-1.5em}

\begin{align*}
    R_{vel}(s_t) &= \frac{1}{N} \sum_{i=1}^{N}{v^i_t} \\
    R_{mpg}(s_t) &= \frac{1}{1609 N} \sum_{i=1}^{N}{\frac{v^i_t}{g^i_t}} \\
    R_{comb}(s_t) &= \alpha R_{vel}(s_t) + \beta R_{mpg}(s_t) - \lambda \|a_t - a_{t-1}\|_{2} \\
\end{align*}
\vspace*{-3em}

\begin{table*}[ht!]
  \centering
  \caption{\textbf{Driving Performance Benchmark}.
    % We show through experiments that our traffic-aware module for extra supervision not only benefits the quality of driving for the entire traffic state, but also for the individual autonomous vehicle. 
    Arrows denote the direction of improvement. We benchmark two baselines, LBC~\cite{chen2019lbc} and TransFuser~\cite{Chitta2022PAMI, Prakash2021CVPR}, and evaluate our method on CARLA Leaderboard Test Scenarios and Longest6 benchmarks from~\cite{Prakash2021CVPR} respectively. Our method, TrAAD, is able to complement and enhance the driving performance of existing methods solely through the distillation of acceleration behaviors, with no modifications to design governing the learning of other controls. For TransFuser, ours is able to improve the RouteCompletion \%, yet incurring negligible infractions.  See Section~\ref{sec:results} for details.} 
    \label{table:benchmark} 
  \begin{tabular}{lcccccccc}
    \toprule
    Method  & DrivingScore$\uparrow$ & RouteCompletion$\uparrow$ & PedCollisions$\downarrow$ & VehCollisions$\downarrow$ & OtherCollisions$\downarrow$ & Timeouts$\downarrow$\\
    \midrule
    LBC & 2.876 & 6.292 & 0.0 & 4.148 & 25.715 & 0.830  \\
    LBC+Ours & \textbf{6.256} & \textbf{19.487} & 0.0 & \textbf{3.706} & \textbf{12.766} & \textbf{0.137}  \\
    \midrule
    TransFuser & \textbf{33.908} & 77.657 & 0.0240 & 3.366 & \textbf{0.168} & 0.144\\
    TransFuser+Ours & 31.291 & \textbf{79.964} & \textbf{0.0223} & \textbf{2.604} & 0.200 & \textbf{0.044}\\
    \bottomrule
  \end{tabular}
  % \vspace*{-0.5em}
\end{table*}

\subsection{Phase 2: Learn to do Everything Else}
This phase involves the integration of the learned Phase 1 acceleration model into an autonomous driving framework for single-vehicle control. In our experiments, we use Learning by Cheating (LBC)~\cite{chen2019lbc} as the backbone for Phase 2. We first train the privileged agent that has access to a top-down ground truth map of the environment $M \in \{0,1\}^{W\times H\times7}$. The privileged agent learns to drive an AV in a fully-supervised manner, similar to LBC. 
Our Phase 1 model, illustrated in the top half of Figure~\ref{fig:method}, directly provides ground truth for the supervision of throttle control in this step. Instead of learning the throttle based on ground-truth labels provided by the expert agent during data collection, the Phase 1 acceleration agent supervises the throttle command. 
Each sample is also annotated with the local traffic lane state, which is also provided to the Phase 1 model. 

Intuitively, the Phase 1 model is teaching the Phase 2 model how to accelerate; all other controls are learned through the original LBC pipeline. This includes the non-cheating ``student'' model, which learns from limited single-view information. Supervision of both speed and steering of the student model is done by the trained Phase 2 privileged model, which should have learned optimized acceleration behavior from the Phase 1 acceleration model. 

This implementation design provides two main advantages: (1) Phases 1 and 2 are modular, and thus can be re-used to train multiple Phase 2 models; (2) Since traffic information transitions between models as a form of knowledge distillation, TrAAD is able to complement prior works rather than competing against them. The involvement of traffic information in our method can be thought of as knowledge distillation.

\begin{table*}[ht!]
  \centering
  \caption{\textbf{Ablation on Impact of Each Factor in Reward Function}. % We show results for several experiments where one reward term is ablated to show the effect of each on societal and individual performance metrics. 
  Arrows denote the direction of improvement. 
  As expected, the model omitting the fuel consumption term is able to freely accelerate without worrying about fuel efficiency. Despite fuel efficiency and traffic flow being inversely related, our model is able to achieve a middle ground between ablation models for the best of both worlds. More details on this experiment can be found in Section~\ref{sec:results}.} 
  \label{table:ablation} 
  \begin{tabular}{lccccccccc}
    \toprule
    % \multicolumn{6}{c}{Real World}{Simulated}                   \\
    % \cmidrule(r){1-2}
    % Jerk ($\Delta m/s$)
    Method  & TrafficFlow$\uparrow$ & FuelCons$\downarrow$ & Jerk$\downarrow$ & DrivingScore$\uparrow$ & RtCompletion$\uparrow$ & Infractions$\downarrow$ & Collisions$_Veh$ $\downarrow$ & Collisions$_{Other}$ $\downarrow$  \\
    \midrule
    Baseline & 1.713 & 1.0399 & 2.2862e-3 & 4.021 & 6.184 & 0.302 & 2.879 & 10.557 \\ 
    NoVelocity & 2.637 & 1.1010 & 1.7311e-3 & 4.197 & 6.871 & 0.350 & 2.309 & 15.701 \\ 
    NoFuel & \textbf{2.679} & 1.0866 & 0.9465e-3 & \textbf{5.507} & \textbf{7.622} & 0.25 & 0.819 & 11.05  \\ 
    NoJerk & 2.290 & \textbf{1.0248} & 0.9466e-3 & 3.900 & 5.263 & \textbf{0.229} & \textbf{0.608} & 9.731 \\
    Ours & 2.428 & 1.0594 & \textbf{0.9465e-3} & 4.094 & 6.271 & 0.320 & 2.801 & \textbf{9.801} \\ 
    \bottomrule 
  \end{tabular} 
  \vspace*{-1.5em}
\end{table*}

% \vspace*{-1.5em}
\subsection{Improving Sample Efficiency with Traffic Gradients}
\label{sec:sample_enhancement}
Since agent-based forward simulation of traffic can be represented by ODEs, gradients of the traffic state at timestep $t$  with respect to that at timestep $t-1$ can be analytically derived. We define these gradients as ``traffic gradients". 
We can use traffic gradients in enhancing the sample efficiency of the common on-policy RL algorithms, such as PPO~\cite{schulman2017proximal} or TRPO~\cite{schulman2015high}. Inspired by the sampling enhancement scheme of \cite{qiao2021efficient}, which applies the scheme to a model-based RL algorithm, here we present a method that is applicable to the other general on-policy RL algorithms and show its efficacy in the Section~\ref{sec:results}

Specifically, PPO and TRPO are based on evaluating the perturbed policy $\widetilde{\pi}$ with the experience from the original policy $\pi$~\cite{schulman2015trust}. This is possible because the expected reward of the policy $\widetilde{\pi}$ can be approximated with the information from the original policy up to the first order. Our strategy is based on manipulating the collected set of experience with our gradients so that we can maximize the efficiency of each update. 

Let us denote a single experience unit as $(s_n, a_n, r_n, s_{n+1}, \frac{\partial r_n}{\partial a_n}, \frac{\partial s_{n+1}}{\partial a_n})$, where $s_n$, $a_n$, and $r_n$ refer to the state, action, and reward at the time step $n$, and $\frac{\partial r_n}{\partial a_n}$ and $\frac{\partial s_{n+1}}{\partial a_n}$ stand for the gradient of the reward and the next state with respect to the action. These gradient terms come from our differentiable traffic simulator. Then we can perturb $(a_n, r_n, s_{n+1})$ with the gradient terms as follows:

\vspace*{-1.5em}
 \begin{align*}
     \widetilde{a}_n = a_n + \epsilon, \quad
     &\widetilde{r}_n = r_n + \epsilon \cdot \frac{\partial r_n}{\partial a_n}, \quad
     \widetilde{s}_{n+1} = s_n + \epsilon \cdot \frac{\partial s_n}{\partial a_n} \\
     &\text{where } |\epsilon \cdot \frac{\partial s_n}{\partial a_n}| \le \delta \text{ for some } \delta > 0.
 \end{align*}
 \vspace*{-1.5em}
 
 Note that the amount of perturbation $\epsilon$ is bounded by a threshold factor $\delta$, as we do not want to perturb $s_{n+1}$ too much. This is because we still want to take advantage of the commonly used advanced advantage estimation techniques, such as GAE~\cite{schulman2015high}, in estimating the advantage, and they usually consider multiple time steps in a trajectory. With this constraint, we ensure that the perturbed trajectory does not deviate from the original one too much so that we can still use the advantage estimation techniques that use the whole trajectory. 
 
 By perturbing the action, reward, and next state in this fashion, we can expect that our experience buffer is filled with more ``important" experience than before, relative to the reward function. Intuitively, an action that produces a higher immediate reward than other actions is more likely to bring about more meaningful results. This intuition is not true in every case, but in many cases, we can observe that it holds. By using this more important experience and manipulating our policy based on it, we can expect our policy to learn more meaningful lessons from the same experience than before. 
 In experiments (Figure~\ref{fig:grad_rl}), gradient-enhanced algorithms are prefixed with ``Diff''. Thus, gradient-enhanced TRPO becomes DiffTRPO, and gradient-enhanced PPO becomes DiffPPO.

%% file: sections/results.tex
\section{Implementation and Results}
\label{sec:results}
% \subsection*{Improving Traffic Flow}

We show results for improved traffic flow via \href{https://leaderboard.carla.org/scenarios/}{CARLA Leaderboard test scenarios}, which describe 10 categories of scenarios based on the National Highway Traffic Safety Administration (NHTSA) pre-crash typology. For experiments with TransFuser in Table~\ref{table:benchmark}, we use the Longest6 benchmark presented in~\cite{Chitta2022PAMI}.

\subsection{Integration with SOTA and Improvements}
\label{subsec:sota}
We implement TrAAD on recent state-of-the-art methods for imitation learning to demonstrate its generalizability and benefits on higher-performing benchmarks. 
Evaluation of learned driving policies for each model is based on CARLA driving benchmark metrics, which quantify route completion, infractions, collisions, and timeouts, among others.
We observe from the top half of the results in Table~\ref{table:benchmark} that TrAAD implemented on Learning By Cheating (LBC)~\cite{chen2019lbc} is able to improve Route Completion (\%) of scenario route driven, by a factor of {\it over 2x}. In addition, we achieve lower overall collisions with pedestrians, other vehicles, and other objects. Agent timeouts are also improved by a factor of 6. Overall, we improve the overall Driving Score by {\it over 2x} as well, where Driving Score is defined as the product between the Route Completion and an infraction penalty. 

Mean traffic flow and mean fuel consumption for each benchmark CARLA scenario are visualized in Figure~\ref{fig:barchart}. Our model consistently improves overall traffic flow over baseline models in the local lane, while maintaining similar fuel consumption despite the natural relation between higher velocities and higher fuel consumption. 
% Note that Scenario 8 was a failure case for our model, while Scenario 25 was a failure case of the baseline model. 

We also evaluate TrAAD on the most recent state-of-the-art model from 2022, TransFuser~\cite{Prakash2021CVPR}. While this method was originally published in 2021, we use the improved 2022 version of the method. Since this method achieves significantly better performance than LBC, we can observe the scaling effects of our method on a driving model with higher benchmark metrics. Despite improving Route Completion and collisions against pedestrians and other vehicles, we observe a slightly worse Driving Score. This is most likely due to our model incurring a marginally higher rate of traffic infractions. Since TrAAD increases overall flow, a slightly higher rate of infractions is likely as a vehicle may be traveling at a higher velocity when encountering red lights or stop signs. However, this slight degradation is minimal and does not actually result in more collisions overall. 

\vspace*{-0.5em}
\subsection{Ablation: Effect of Traffic Gradients on Performance}
We show the impact of traffic gradients and perturbation threshold values $\delta$ during training for the on-policy algorithm TRPO~\cite{Schulman_Levine_Abbeel_Jordan_Moritz_07} in Figure~\ref{fig:grad_rl}. In this experiment, we compare the training reward curve of TRPO versus gradient-enhanced DiffTRPO with different perturbation threshold $\delta$ values. 
We do not evaluate this enhancement on CARLA benchmark scenarios, as not every scenario involves a difficult traffic-related task. If the task is not difficult enough, the benefits of using traffic gradients are less pronounced in learning. To demonstrate improvement in difficult traffic tasks, we use the Figure-Eight benchmark from~\cite{wu2017flow, pmlr-v87-vinitsky18a}. 
In this environment, the ego vehicle is tasked with controlling dense traffic flow to maximize overall traffic flow to a target speed, i.e. {\it congestion and shockwaves are undesirable in the optimal policy}. 

While all DiffTRPO iterations visibly perform better than the baseline TRPO, threshold values of $\delta=0.2$ and $\delta=0.4$ converge the fastest to the highest reward. 
Each training curve is averaged over ten runs to account for stochasticity. 
More results for PPO can be found in supplemental materials, where $\delta=0.1$ achieves the best results for DiffPPO. 

\subsection{Ablation: Effect of Each Reward Term}
We study the effect of each of the three terms of the reward function. In this ablation experiment, we analyze three scenarios: (1) Without velocity term, (2) Without fuel consumption term, (3) Without jerkiness constraint. We evaluate them on societal metrics of traffic flow, fuel consumption, and driving jerkiness, as well as evaluation metrics from CARLA driving benchmark similar to Section~\ref{subsec:sota}. Results for this experiment can be found in Table~\ref{table:ablation}. 

Overall, we find that our model expectedly achieves a middle ground between ablated models. In our combined reward function, we optimize for both fuel efficiency and traffic flow. However, these terms are inversely related in the real world; as overall traffic flow increases, more fuel is used in order to produce a higher velocity. Thus, we observe from the results that the model omitting the fuel term is able to achieve the highest overall traffic flow, as fuel efficiency does not constrain the optimization of traffic flow. In addition, omitting the jerk constraint results in fewer infractions. For scenarios involving yielding to traffic laws, such as red lights and stop signs, restricting jerkiness may harm infraction scores as vehicles cannot suddenly decelerate to a complete stop. Overall, we observe that TrAAD achieves the ``best of both worlds'' across {\em all} other metrics, while the ablated models achieve the best results solely on individual metrics.

% \vspace*{-1em}

%% file: sections/conclusion.tex
% \vspace*{-0.5em}
\section{Conclusion and Discussion}
% \vspace*{-0.5em}

In this paper, we present a method for traffic-aware autonomous driving (TrAAD) that benefits both the autonomous vehicle and the traffic around it. TrAAD complements existing work by providing additional supervision on acceleration behavior guided by differentiable traffic simulation. In addition, we improve the sample efficiency of on-policy RL algorithms using analytical gradients of car-following models and show that our method produces a driving policy that not only benefits the surrounding traffic system but also improves the driving performance of each autonomous vehicle on multiple benchmarks. Distillation results also imply interesting future work for distillation for real-world models. 

\subsection*{Limitations and Future Work}  Independent of our work, the current implementation integrated with CARLA simulation for learning imposes some limitations.  For example, the shortage of difficult traffic scenarios on available driving benchmarks within CARLA limits more evaluation of our approach with FLOW on vehicle control; similarly with limited traffic tasks on the current driving benchmarks. Furthermore, current driving benchmarks do not consider societal goals, such as the overall traffic flow and fuel consumption, and thus may downplay the potential positive impact of our method.  
More driving benchmarks that consider the global objectives of the entire road network system would allow for a more comprehensive evaluation of learned policies for AVs.

%% file: sections/appendix.tex
\onecolumn
\section*{APPENDIX}
\subsection{Implementation Details}

\subsubsection{TransFuser Experiments}
We recollect the TransFuser dataset by using our co-simulation wrapper to record corresponding traffic information for each sample. We then train our method on our dataset via transfer learning, where we use pre-trained weights provided by TransFuser authors. Each TransFuser model provides driving control prediction via an ensemble of three trained models from different initialization seeds. Our model and the baseline are initialized with the same three initialization weights. 

\subsection{Formal Definition of the Intelligent Driver Model (IDM)}

Let $x_\alpha$ and $v_\alpha$ be the position and velocity of a vehicle $\alpha$ in line with the leading vehicle for a fixed time step. Note that $\alpha - 1$ denotes the position of the vehicle in front, so $\alpha = 0$ for the leading vehicle.
{\small
\begin{align*}
    {\dot {x}}_{\alpha }&={\frac {\mathrm {d} x_{\alpha }}{\mathrm {d} t}}=v_{\alpha } \\
    {\dot {v}}_{\alpha }&={\frac {\mathrm {d} v_{\alpha }}{\mathrm {d} t}}=a\,\left(1-\left({\frac {v_{\alpha }}{v_{0}}}\right)^{\delta }-\left({\frac {s^{*}(v_{\alpha },\Delta v_{\alpha })}{s_{\alpha }}}\right)^{2}\right) \\
    s_\alpha &= x_{\alpha-1} - x_\alpha - l_{\alpha-1} \\
    s^{*}(v_{\alpha },\Delta v_{\alpha })&=s_{0}+v_{\alpha }\,T+{\frac {v_{\alpha}\,\Delta v_{\alpha }}{2\,{\sqrt {a\,b}}}} \\
    \Delta v_\alpha &= v_\alpha - v_{\alpha-1}
\end{align*}
}%
    
\noindent

% \subsection{IDM Variables}
\begin{table}[h!]
\centering
\begin{tabular}{l|l}
\textbf{Variable} & \textbf{Definition} \\
\hline
& \\
$v_0$ & Desired velocity \\
$T$ & Safe time headway \\
$a$ & Maximum acceleration \\
$b$ & Comfortable deceleration \\
$\delta$ & Minimum distance \\
$l$ & Vehicle length \\
$\Delta v_\alpha$ & Velocity difference with the front vehicle
\end{tabular}
\caption{\label{tab:def}Variable descriptions for the Intelligent Driver Model.}
\end{table}

\subsection{Jacobian of the IDM}
\label{subsec:jacobian}
For ease of calculation, we represent the simulation state vector $q$ at a certain time step $t$ to be a $(1,2N)$ vector instead of a $(N,2)$ vector, where $N$ is the number of vehicles. Thus, the simulation state will take on the form

\begin{align*}
    q_t = 
    \begin{bmatrix}
    x_1 & v_1 & ... & x_N & v_N
    \end{bmatrix}
\end{align*}

Then, we can expect the Jacobian relating one state to the next to be a vector of dimension $(2N, 2N)$. For a particular vehicle $\alpha$ indices from the front vehicle, the Jacobian of the IDM forward simulation is derived. Let 

% \begin{align*}
%     \frac{dx_\alpha}{dt} &= f(x_\alpha, v_\alpha) \\
%     \frac{dv_\alpha}{dt} &= g(x_\alpha, v_\alpha) 
% \end{align*}
\begin{align*}
    f(x_\alpha, v_\alpha) &= {\dot {x}}_{\alpha } \\    
    g(x_\alpha, v_\alpha) &= {\dot {v}}_{\alpha } 
\end{align*}

Then the Jacobian of the IDM with respect to state values position $x$ and velocity $v$ will take on the form: 

\begin{align*}
    J_{idm}(q_0, q_1) &= 
    \begin{pmatrix}
    \frac{\delta vehicle1_{t=1}}{\delta vehicle1_{t=0}} & ... & \frac{\delta vehicle1_{t=1}}{\delta vehicleN_{t=0}} \\
    \vdots & \ddots & \vdots \\ 
    \frac{\delta vehicleN_{t=1}}{\delta vehicle1_{t=0}} & ... & \frac{\delta vehicleN_{t=1}}{\delta vehicleN_{t=0}}
    \end{pmatrix} \\
\end{align*}

Recall from Equation~\ref{eq:sim_state} that the state of a single agent or vehicle comprises both a position and a velocity component. For sake of readability, the derivative below is always taken with respect to the previous timestep. Then, the Jacobian above can then be expanded:

% \begin{align*}
%     % J_{idm}(q_0, q_1) &= 
%     \begin{pmatrix}
%     \frac{\partial f(x_1,v_1)}{\partial x_1} & \frac{\partial f(x_1,v_1)}{\partial v_1} & ... & ... & \frac{\partial f(x_1,v_1)}{\partial x_N} & \frac{\partial f(x_1,v_1)}{\partial v_N} \\ 
%     \frac{\partial g(x_1,v_1)}{\partial x_1} & \frac{\partial g(x_1,v_1)}{\partial v_1} & ... & ... & \frac{\partial g(x_1,v_1)}{\partial x_N} & \frac{\partial g(x_1,v_1)}{\partial v_N} \\ 
%     \vdots & \vdots & \ddots & \ddots & \vdots & \vdots \\
%     \vdots & \vdots & \ddots & \ddots & \vdots & \vdots \\
%     \frac{\partial f(x_N,v_N)}{\partial x_1} & \frac{\partial f(x_N,v_N)}{\partial v_1} & ... & ... & \frac{\partial f(x_N,v_N)}{\partial x_N} & \frac{\partial f(x_N,v_N)}{\partial v_N} \\ 
%     \frac{\partial g(x_N,v_N)}{\partial x_1} & \frac{\partial g(x_N,v_N)}{\partial v_1} & ... & ... & \frac{\partial g(x_N,v_N)}{\partial x_N} & \frac{\partial g(x_N,v_N)}{\partial v_N} \\ 
%     \end{pmatrix} \\
% \end{align*}

\begin{align*}
    % J_{idm}(q_0, q_1) &= 
    \begin{pmatrix}
    \frac{\partial f(x_1,v_1)}{\partial x_1} & \frac{\partial f(x_1,v_1)}{\partial v_1} & ... & ... & \frac{\partial f(x_1,v_1)}{\partial x_N} & \frac{\partial f(x_1,v_1)}{\partial v_N} \\ 
    \frac{\partial g(x_1,v_1)}{\partial x_1} & \frac{\partial g(x_1,v_1)}{\partial v_1} & ... & ... & \frac{\partial g(x_1,v_1)}{\partial x_N} & \frac{\partial g(x_1,v_1)}{\partial v_N} \\ 
    \vdots & \vdots & \ddots & \ddots & \vdots & \vdots \\
    \vdots & \vdots & \ddots & \ddots & \vdots & \vdots \\
    \frac{\partial f(x_N,v_N)}{\partial x_1} & \frac{\partial f(x_N,v_N)}{\partial v_1} & ... & ... & \frac{\partial f(x_N,v_N)}{\partial x_N} & \frac{\partial f(x_N,v_N)}{\partial v_N} \\ 
    \frac{\partial g(x_N,v_N)}{\partial x_1} & \frac{\partial g(x_N,v_N)}{\partial v_1} & ... & ... & \frac{\partial g(x_N,v_N)}{\partial x_N} & \frac{\partial g(x_N,v_N)}{\partial v_N} \\ 
    \end{pmatrix} \\
\end{align*}

This resulting Jacobian ends up being a lower triangular matrix. This is because any entries above the main 2-by-2 diagonal represent the relation between a vehicle and the vehicles behind it. In car-following models, vehicle position and velocity are not affected by vehicles behind. Thus, the upper half of the Jacobian is zeroed out. Additionally, in the context of IDM, a vehicle is only influenced by the vehicle directly in front of it. Thus, any partial derivatives between a vehicle and any vehicle more than 1 position ahead are also zeroed out. An example of the Jacobian for a 3-vehicle simulation is shown below:

\begin{align*}
    % J_{idm}(q_0, q_1) &= 
    \begin{pmatrix}
    \frac{\partial f(x_1,v_1)}{\partial x_1} & \frac{\partial f(x_1,v_1)}{\partial v_1} & \textbf{0} & \textbf{0} & \textbf{0} & \textbf{0}\\ 
    \frac{\partial g(x_1,v_1)}{\partial x_1} & \frac{\partial g(x_1,v_1)}{\partial v_1} & \textbf{0} & \textbf{0} & \textbf{0} & \textbf{0}\\ 
    \frac{\partial f(x_2,v_2)}{\partial x_1} & \frac{\partial f(x_2,v_2)}{\partial v_1} & \frac{\partial f(x_2,v_2)}{\partial x_2} & \frac{\partial f(x_2,v_2)}{\partial v_2} & \textbf{0} & \textbf{0} \\
    \frac{\partial g(x_2,v_2)}{\partial x_1} & \frac{\partial g(x_2,v_2)}{\partial v_1} & \frac{\partial g(x_2,v_2)}{\partial x_2} & \frac{\partial g(x_2,v_2)}{\partial v_2} & \textbf{0} & \textbf{0} \\
    \textbf{0} & \textbf{0} & \frac{\partial f(x_3,v_3)}{\partial x_2} & \frac{\partial f(x_3,v_3)}{\partial v_2} & \frac{\partial f(x_3,v_3)}{\partial x_3} & \frac{\partial f(x_3,v_3)}{\partial v_3} \\ 
    \textbf{0} & \textbf{0} & \frac{\partial g(x_3,v_3)}{\partial x_2} & \frac{\partial g(x_3,v_3)}{\partial v_2} & \frac{\partial g(x_3,v_3)}{\partial x_3} & \frac{\partial g(x_3,v_3)}{\partial v_3} \\ 
    \end{pmatrix} \\
\end{align*}

From this example, we can see that the Jacobian can be found by generalizing the partial derivative on the 2-by-2 diagonal, as well as the partial derivatives one 2-by-2 row directly below the diagonal. There are thus eight generalized terms to build the Jacobian matrix of one simulation time step. One entry on the 2-by-2 diagonal of the Jacobian matrix can be intuitively defined as the Jacobian of a vehicle's state with respect to itself from the previous time step. Thus, every 2-by-2 entry on the diagonal takes this general form for a particular vehicle in index $\alpha$ from the front: 

{\small
\begin{align*}
    J_{idm}[2\alpha,2\alpha] &= 
    \begin{pmatrix}
        \frac{\partial f}{\partial x_\alpha} & \frac{\partial f}{\partial v_\alpha} \\
        \frac{\partial g}{\partial x_\alpha} & \frac{\partial g}{\partial v_\alpha}
    \end{pmatrix} \\
    \frac{\partial f}{\partial x_\alpha} &= 0 \\
    \frac{\partial f}{\partial v_\alpha} &= 1 \\
    \frac{\partial g}{\partial x_\alpha} &= \frac{-2as^*(v_\alpha, \Delta v_\alpha)^2}{s_\alpha^3} \\
    \frac{\partial g}{\partial v_\alpha} &= \frac{-a \delta v_\alpha^{\delta-1}}{v_0^\delta} + \frac{-2a}{s_\alpha^2}\left(T + \frac{\Delta v_\alpha + v_\alpha}{2\sqrt{ab}}\right)s^*(v_\alpha, \Delta v_\alpha)
\end{align*}
}%

And, likewise, the diagonal one 2-by-2 row directly beneath it will take the form (excluding the first row, which represents the leading vehicle): 

\begin{align*}
    J_{idm}[2\alpha,2\alpha - 2] &= 
    \begin{pmatrix}
        \frac{\partial f}{\partial x_{\alpha-1}} & \frac{\partial f}{\partial v_{\alpha-1}} \\
        \frac{\partial g}{\partial x_{\alpha-1}} & \frac{\partial g}{\partial v_{\alpha-1}}
    \end{pmatrix} \\
    \frac{\partial f}{\partial x_{\alpha-1}} &= 0 \\
    \frac{\partial f}{\partial v_{\alpha-1}} &= 0 \\
    \frac{\partial g}{\partial x_{\alpha-1}} &= \frac{2as^*(v_\alpha, \Delta v_\alpha)^2}{s_\alpha^3} \\
    \frac{\partial g}{\partial v_{\alpha-1}} &= \frac{2as^*(v_\alpha, \Delta v_\alpha)v_\alpha}{2s_\alpha^2\sqrt{ab}}
\end{align*}

With this analytically derived form of the Jacobian, we can now compute the Jacobian at any time step directly without the use of Autograd, given the current state as input. For $N$ vehicles, we calculate $4N$ values on the main diagonal, $4(N-1)$ values on the "sub-diagonal", and zero out every other value to attain the $2N \times 2N$ Jacobian matrix for a particular time step.

Note that sometimes, IDM will produce negative velocities in intermediate or end states. To address this issue, any submatrices where IDM yields a negative value at the next time step will have a gradient of zeroes. This addresses negative velocity clipping in the forward simulation, where negative velocities after the simulation update are clipped to zero. 

\begin{figure}[htb!]
  \centering
  \includegraphics[width=8cm,height=6cm]{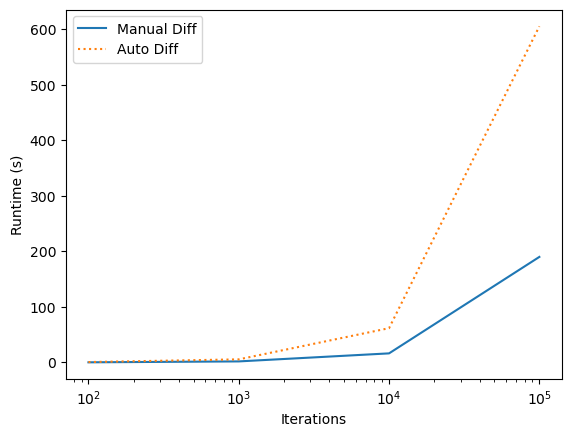}
  \caption{Comparison on runtime (s) over the number of iterations for analytical differentiation versus auto differentiation of the car-following model IDM. }
  \label{fig:runtime}
\end{figure}

The boost in speed per iteration using the analytical gradient versus auto-differentiation is visualized in Figure~\ref{fig:runtime}.

%% file: root.bbl
\begin{thebibliography}{10}
\providecommand{\url}[1]{#1}
\csname url@rmstyle\endcsname
\providecommand{\newblock}{\relax}
\providecommand{\bibinfo}[2]{#2}
\providecommand\BIBentrySTDinterwordspacing{\spaceskip=0pt\relax}
\providecommand\BIBentryALTinterwordstretchfactor{4}
\providecommand\BIBentryALTinterwordspacing{\spaceskip=\fontdimen2\font plus
\BIBentryALTinterwordstretchfactor\fontdimen3\font minus
  \fontdimen4\font\relax}
\providecommand\BIBforeignlanguage[2]{{%
\expandafter\ifx\csname l@#1\endcsname\relax
\typeout{** WARNING: IEEEtran.bst: No hyphenation pattern has been}%
\typeout{** loaded for the language `#1'. Using the pattern for}%
\typeout{** the default language instead.}%
\else
\language=\csname l@#1\endcsname
\fi
#2}}

\bibitem{emissions}
\BIBentryALTinterwordspacing
J.~Kim, ``Estimating the social cost of congestion using the bottleneck
  model,'' Mar 2019. [Online]. Available:
  \url{https://papers.ssrn.com/abstract=3356167}
\BIBentrySTDinterwordspacing

\bibitem{EPA_emissions}
``Sources of greenhouse gas emissions,''
  \url{https://www.epa.gov/ghgemissions/sources-greenhouse-gas-emissions},
  accessed: 2022-09.

\bibitem{CDC_cause_of_death}
``Road traffic injuries and deaths—a global problem,''
  \url{https://www.cdc.gov/injury/features/global-road-safety/index.html},
  accessed: 2022-09.

\bibitem{Treiber_Hennecke_Helbing_2000}
M.~Treiber, A.~Hennecke, and D.~Helbing, ``Congested traffic states in
  empirical observations and microscopic simulations,'' \emph{Physical Review
  E}, vol.~62, no.~2, p. 1805–1824, Aug 2000.

\bibitem{Newell_2002}
G.~F. Newell, ``A simplified car-following theory: a lower order model,''
  \emph{Transportation Research Part B: Methodological}, vol.~36, no.~3, p.
  195–205, Mar 2002.

\bibitem{Wiedemann_1974}
R.~Wiedemann, \emph{\BIBforeignlanguage{de}{Simulation des
  Strassenverkehrsflusses}}.\hskip 1em plus 0.5em minus 0.4em\relax Karlsruhe:
  Institut für Verkehrswesen der Universität Karlsruhe, 1974.

\bibitem{krauss}
S.~Krauss, ``Microscopic modeling of traffic flow: investigation of collision
  free vehicle dynamics,'' Apr 1998.

\bibitem{lighthill1955kinematic}
M.~J. Lighthill and G.~B. Whitham, ``On kinematic waves ii. a theory of traffic
  flow on long crowded roads,'' \emph{Proceedings of the Royal Society of
  London. Series A. Mathematical and Physical Sciences}, vol. 229, no. 1178,
  pp. 317--345, 1955.

\bibitem{gazis1961nonlinear}
D.~C. Gazis, R.~Herman, and R.~W. Rothery, ``Nonlinear follow-the-leader models
  of traffic flow,'' \emph{Operations research}, vol.~9, no.~4, pp. 545--567,
  1961.

\bibitem{aw2000resurrection}
A.~Aw and M.~Rascle, ``Resurrection of" second order" models of traffic flow,''
  \emph{SIAM journal on applied mathematics}, vol.~60, no.~3, pp. 916--938,
  2000.

\bibitem{zhang2002non}
H.~M. Zhang, ``A non-equilibrium traffic model devoid of gas-like behavior,''
  \emph{Transportation Research Part B: Methodological}, vol.~36, no.~3, pp.
  275--290, 2002.

\bibitem{Kerner}
B.~S. Kerner, \emph{Introduction to Modern Traffic Flow Theory and
  Control}.\hskip 1em plus 0.5em minus 0.4em\relax Springer Berlin Heidelberg.

\bibitem{Olayode_Tartibu_Okwu_2021}
I.~O. Olayode, L.~K. Tartibu, and M.~O. Okwu, ``Prediction and modeling of
  traffic flow of human-driven vehicles at a signalized road intersection using
  artificial neural network model: A south african road transportation system
  scenario,'' \emph{Periodica Polytechnica: Transportation Engineering},
  vol.~6, p. 100095, Dec 2021.

\bibitem{Diehl_Brunner_Le_Knoll_2019}
F.~Diehl, T.~Brunner, M.~T. Le, and A.~Knoll, ``Graph neural networks for
  modelling traffic participant interaction,'' in \emph{2019 IEEE Intelligent
  Vehicles Symposium (IV)}.\hskip 1em plus 0.5em minus 0.4em\relax
  ieeexplore.ieee.org, Jun 2019, p. 695–701.

\bibitem{Jiang_Luo_2022}
W.~Jiang and J.~Luo, ``Graph neural network for traffic forecasting: A
  survey,'' \emph{Expert systems with applications}, vol. 207, p. 117921, Nov
  2022.

\bibitem{Do_Taherifar_Vu_2019}
L.~N.~N. Do, N.~Taherifar, and H.~L. Vu, ``\BIBforeignlanguage{en}{Survey of
  neural network‐based models for short‐term traffic state prediction},''
  \emph{\BIBforeignlanguage{en}{Wiley interdisciplinary reviews. Data mining
  and knowledge discovery}}, vol.~9, no.~1, p. e1285, Jan 2019.

\bibitem{Zhu_Wang_Pu_Hu_Wang_Ke_2020}
M.~Zhu, Y.~Wang, Z.~Pu, J.~Hu, X.~Wang, and R.~Ke, ``Safe, efficient, and
  comfortable velocity control based on reinforcement learning for autonomous
  driving,'' \emph{Transportation Research Part C: Emerging Technologies}, vol.
  117, p. 102662, Aug 2020.

\bibitem{Shen_Zhang_Ouyang_Li_Raksincharoensak_2020}
X.~Shen, X.~Zhang, T.~Ouyang, Y.~Li, and P.~Raksincharoensak, ``Cooperative
  comfortable-driving at signalized intersections for connected and automated
  vehicles,'' \emph{IEEE Robotics and Automation Letters}, vol.~5, no.~4, p.
  6247–6254, Oct 2020.

\bibitem{Zhu_Wang_Hu_2020}
M.~Zhu, X.~Wang, and J.~Hu, ``Impact on car following behavior of a forward
  collision warning system with headway monitoring,'' \emph{Transportation
  Research Part C: Emerging Technologies}, vol. 111, p. 226–244, Feb 2020.

\bibitem{Li_Yang_Li_Qu_Lyu_Li_2022}
G.~Li, Y.~Yang, S.~Li, X.~Qu, N.~Lyu, and S.~E. Li, ``Decision making of
  autonomous vehicles in lane change scenarios: Deep reinforcement learning
  approaches with risk awareness,'' \emph{Transportation Research Part C:
  Emerging Technologies}, vol. 134, p. 103452, Jan 2022.

\bibitem{Wegener_Koch_Eisenbarth_Andert_2021}
M.~Wegener, L.~Koch, M.~Eisenbarth, and J.~Andert, ``Automated eco-driving in
  urban scenarios using deep reinforcement learning,'' \emph{Transportation
  Research Part C: Emerging Technologies}, vol. 126, p. 102967, May 2021.

\bibitem{Andelfinger_2021}
\BIBentryALTinterwordspacing
P.~Andelfinger, ``Differentiable agent-based simulation for gradient-guided
  simulation-based optimization,'' in \emph{Proceedings of the 2021 ACM SIGSIM
  Conference on Principles of Advanced Discrete Simulation}.\hskip 1em plus
  0.5em minus 0.4em\relax New York, NY, USA: ACM, May 2021. [Online].
  Available: \url{https://dl.acm.org/doi/10.1145/3437959.3459261}
\BIBentrySTDinterwordspacing

\bibitem{son2022traffic}
\BIBentryALTinterwordspacing
S.~Son, Y.-L. Qiao, J.~Sewall, and M.~C. Lin, ``Differentiable hybrid traffic
  simulation,'' \emph{ACM Trans. Graph.}, vol.~41, no.~6, nov 2022. [Online].
  Available: \url{https://doi.org/10.1145/3550454.3555492}
\BIBentrySTDinterwordspacing

\bibitem{9210154}
S.~Aradi, ``Survey of deep reinforcement learning for motion planning of
  autonomous vehicles,'' \emph{IEEE Transactions on Intelligent Transportation
  Systems}, vol.~23, no.~2, pp. 740--759, 2022.

\bibitem{Qiao_Muelling_Dolan_Palanisamy_Mudalige_2018}
Z.~Qiao, K.~Muelling, J.~M. Dolan, P.~Palanisamy, and P.~Mudalige,
  ``Automatically generated curriculum based reinforcement learning for
  autonomous vehicles in urban environment,'' in \emph{2018 IEEE Intelligent
  Vehicles Symposium (IV)}, Jun 2018, p. 1233–1238.

\bibitem{Bouton_Nakhaei_Fujimura_Kochenderfer_2019}
M.~Bouton, A.~Nakhaei, K.~Fujimura, and M.~J. Kochenderfer, ``Cooperation-aware
  reinforcement learning for merging in dense traffic,'' in \emph{2019 IEEE
  Intelligent Transportation Systems Conference (ITSC)}, Oct 2019, p.
  3441–3447.

\bibitem{Kaushik_Prasad_Krishna_Ravindran_2018}
M.~Kaushik, V.~Prasad, K.~M. Krishna, and B.~Ravindran, ``Overtaking maneuvers
  in simulated highway driving using deep reinforcement learning,'' in
  \emph{2018 IEEE Intelligent Vehicles Symposium (IV)}, Jun 2018, p.
  1885–1890.

\bibitem{Ferdowsi_Challita_Saad_Mandayam_2018}
A.~Ferdowsi, U.~Challita, W.~Saad, and N.~B. Mandayam, ``Robust deep
  reinforcement learning for security and safety in autonomous vehicle
  systems,'' in \emph{2018 21st International Conference on Intelligent
  Transportation Systems (ITSC)}, Nov 2018, p. 307–312.

\bibitem{Ma_Driggs-Campbell_Kochenderfer_2018}
X.~Ma, K.~Driggs-Campbell, and M.~J. Kochenderfer, ``Improved robustness and
  safety for autonomous vehicle control with adversarial reinforcement
  learning,'' in \emph{2018 IEEE Intelligent Vehicles Symposium (IV)}.\hskip
  1em plus 0.5em minus 0.4em\relax IEEE Press, Jun 2018, p. 1665–1671.

\bibitem{wu2017flow}
C.~Wu, A.~Kreidieh, K.~Parvate, E.~Vinitsky, and A.~M. Bayen, ``Flow:
  Architecture and benchmarking for reinforcement learning in traffic
  control,'' \emph{arXiv preprint arXiv:1710.05465}, vol.~10, 2017.

\bibitem{SUMO2018}
\BIBentryALTinterwordspacing
P.~A. Lopez, M.~Behrisch, L.~Bieker-Walz, J.~Erdmann, Y.-P. Fl{\"o}tter{\"o}d,
  R.~Hilbrich, L.~L{\"u}cken, J.~Rummel, P.~Wagner, and E.~Wie{\ss}ner,
  ``Microscopic traffic simulation using sumo,'' in \emph{The 21st IEEE
  International Conference on Intelligent Transportation Systems}.\hskip 1em
  plus 0.5em minus 0.4em\relax IEEE, 2018. [Online]. Available:
  \url{https://elib.dlr.de/124092/}
\BIBentrySTDinterwordspacing

\bibitem{pmlr-v87-vinitsky18a}
\BIBentryALTinterwordspacing
E.~Vinitsky, A.~Kreidieh, L.~L. Flem, N.~Kheterpal, K.~Jang, C.~Wu, F.~Wu,
  R.~Liaw, E.~Liang, and A.~M. Bayen, ``Benchmarks for reinforcement learning
  in mixed-autonomy traffic,'' in \emph{Proceedings of The 2nd Conference on
  Robot Learning}, ser. Proceedings of Machine Learning Research, A.~Billard,
  A.~Dragan, J.~Peters, and J.~Morimoto, Eds., vol.~87.\hskip 1em plus 0.5em
  minus 0.4em\relax PMLR, 29--31 Oct 2018, pp. 399--409. [Online]. Available:
  \url{https://proceedings.mlr.press/v87/vinitsky18a.html}
\BIBentrySTDinterwordspacing

\bibitem{Schulman_Levine_Abbeel_Jordan_Moritz_07}
J.~Schulman, S.~Levine, P.~Abbeel, M.~Jordan, and P.~Moritz, ``Trust region
  policy optimization,'' in \emph{Proceedings of the 32nd International
  Conference on Machine Learning}, ser. Proceedings of Machine Learning
  Research, F.~Bach and D.~Blei, Eds., vol.~37.\hskip 1em plus 0.5em minus
  0.4em\relax Lille, France: PMLR, 07--09 Jul 2015, p. 1889–1897.

\bibitem{chen2019lbc}
D.~Chen, B.~Zhou, V.~Koltun, and P.~Kr\"ahenb\"uhl, ``Learning by cheating,''
  in \emph{Conference on Robot Learning (CoRL)}, 2019.

\bibitem{Dosovitskiy17}
A.~Dosovitskiy, G.~Ros, F.~Codevilla, A.~Lopez, and V.~Koltun, ``{CARLA}: {An}
  open urban driving simulator,'' in \emph{Proceedings of the 1st Annual
  Conference on Robot Learning}, 2017, pp. 1--16.

\bibitem{Chitta2022PAMI}
K.~Chitta, A.~Prakash, B.~Jaeger, Z.~Yu, K.~Renz, and A.~Geiger, ``Transfuser:
  Imitation with transformer-based sensor fusion for autonomous driving,''
  \emph{Pattern Analysis and Machine Intelligence (PAMI)}, 2022.

\bibitem{Prakash2021CVPR}
A.~Prakash, K.~Chitta, and A.~Geiger, ``Multi-modal fusion transformer for
  end-to-end autonomous driving,'' in \emph{Conference on Computer Vision and
  Pattern Recognition (CVPR)}, 2021.

\bibitem{schulman2017proximal}
J.~Schulman, F.~Wolski, P.~Dhariwal, A.~Radford, and O.~Klimov, ``Proximal
  policy optimization algorithms,'' \emph{arXiv preprint arXiv:1707.06347},
  2017.

\bibitem{schulman2015high}
J.~Schulman, P.~Moritz, S.~Levine, M.~Jordan, and P.~Abbeel, ``High-dimensional
  continuous control using generalized advantage estimation,'' \emph{arXiv
  preprint arXiv:1506.02438}, 2015.

\bibitem{qiao2021efficient}
Y.-L. Qiao, J.~Liang, V.~Koltun, and M.~C. Lin, ``Efficient differentiable
  simulation of articulated bodies,'' in \emph{International Conference on
  Machine Learning}.\hskip 1em plus 0.5em minus 0.4em\relax PMLR, 2021, pp.
  8661--8671.

\bibitem{schulman2015trust}
J.~Schulman, S.~Levine, P.~Abbeel, M.~Jordan, and P.~Moritz, ``Trust region
  policy optimization,'' in \emph{International conference on machine
  learning}.\hskip 1em plus 0.5em minus 0.4em\relax PMLR, 2015, pp. 1889--1897.

\end{thebibliography}
